    \documentclass{article}
    
    \makeatletter
    \def\input@path{{arxiv/}}
    \makeatother
    \usepackage{iclr2027_conference,times}
    \usepackage[T1]{fontenc}
    \usepackage[utf8]{inputenc}
    \usepackage{graphicx} 
    \graphicspath{{./}{arxiv/}}
    \usepackage{url}
    \usepackage{xcolor}
    
    \usepackage{float}
    \usepackage{wrapfig}
    
    \usepackage{amsmath,amssymb}
    \usepackage{booktabs}
    \usepackage{multirow}
    
    \usepackage{tikz}
    \usepackage{pgfplots}
    \usepgfplotslibrary{groupplots}
    \pgfplotsset{compat=1.18}
    
    \usetikzlibrary{
      positioning,
      fit,
      calc,
      backgrounds,
      arrows.meta,
      decorations.pathreplacing
    }
    
    \definecolor{p1tds}{RGB}{216,112,68}
    \definecolor{myblue}{RGB}{74,119,227}
    \definecolor{extteal}{RGB}{30,150,148}
    \definecolor{embcolor}{RGB}{122,46,62}
    
    \title{Gray-Box Hallucination Detection in LLMs with Token Identities}
    
    \newcommand{\arxivauthor}[4]{%
      \begin{minipage}[t]{#1\linewidth}
        \centering
        \setlength{\parskip}{0pt}%
        \textbf{#2}\par
        \vspace{2pt}%
        \if\relax\detokenize{#3}\relax\else
          {\small\ttfamily #3\par}%
          \vspace{2pt}%
        \fi
        {\small #4\par}%
      \end{minipage}%
    }
    \newcommand{\arxivauthorlist}{%
      \begin{minipage}[t]{\dimexpr\textwidth-2\tabcolsep\relax}
        \normalfont
        \setlength{\parskip}{0pt}%
        \noindent
        \arxivauthor{0.48}{Shaked Goldshtein\textsuperscript{*}}{shakedgoldshtein@gmail.com}{Technion}%
        \hfill
        \arxivauthor{0.48}{Yitzchak Grunbaum\textsuperscript{*}}{itzick.grin@gmail.com}{Technion}%
        \par\vspace{14pt}\noindent
        \arxivauthor{0.32}{Fabrizio Frasca}{}{Technion}%
        \hfill
        \arxivauthor{0.32}{Guy Bar-Shalom}{}{Technion}%
        \hfill
        \arxivauthor{0.32}{Haggai Maron}{}{Technion}%
      \end{minipage}%
    }
    \author{\raisebox{-6pt}{\arxivauthorlist}}
    \iclrfinalcopy
    \usepackage{colortbl}
    \usepackage[hidelinks]{hyperref}
    \hypersetup{
      pdftitle={Gray-Box Hallucination Detection in LLMs with Token Identities},
      pdfauthor={Shaked Goldshtein, Yitzchak Grunbaum, Fabrizio Frasca, Guy Bar-Shalom, Haggai Maron}
    }
    
\begin{document}
    \maketitle
    \fancyhead{}
    \renewcommand{\headrulewidth}{0pt}
    \begingroup
      \renewcommand{\thefootnote}{\ensuremath{\ast}}
      \footnotetext[1]{Equal contribution}
    \endgroup

    \begin{abstract}
    Large language models often produce fluent but factually incorrect output.
    Gray-box detectors identify such hallucinations from the token-probability
    statistics emitted during generation, avoiding the need for white-box access
    to hidden states that frontier models do not provide. These detectors, however, typically rely only on probability-derived representations and ignore
    the identities of the candidate tokens. They can therefore observe how
    uncertain the model is, but not \emph{what it is uncertain about}. Consequently, they
    cannot distinguish between a model that is torn between ``big'' and ``small''
    and one that is torn between ``big'' and 
    ``huge'' when the two pairs receive
    the same probability distribution. We introduce GURI (Gray-box Uncertainty Representations with Identity), a
    lightweight gray-box hallucination detector that combines candidate-token
    probabilities with per-token vector identifiers. This exposes which alternatives the model considered while
    preserving the accessibility and transferability of the gray-box setting. Across three open-weight models and three question-answering benchmarks,
    GURI increases mean test ROC-AUC from \(72.09\%\) to \(79.09\%\)
    over a probability-only baseline. On the closed-source GPT-4.1 model,
    mean ROC-AUC increases from \(67.39\%\) to \(71.84\%\).
  Among the three open-weight models, GURI outperforms transferred
probability-only baseline in 17 of 18 directed cross-model transfers without fine-tuning.
For the three open-weight models, cross-dataset transfer between
HotpotQA and TriviaQA with the model held fixed improves mean
ROC-AUC over transferred LOS-Net by 3.76 points.
    \end{abstract}

    %
    
    \section{Introduction}\label{sec:intro}
    
    Large language models can produce fluent but factually incorrect
    responses, motivating methods for detecting hallucinations
    \citep{ji2023survey,huang2023survey,Maynez2020}.
    White-box detectors use internal model signals, whereas black-box
    methods assess generated text, often using additional model calls.
    Gray-box detection uses token probabilities exposed during generation
    and can operate without internal activations or additional sampling
    \citep{openai_api_logprobs,google_gemini_logprobs}.
    
    Probability-only gray-box methods range from summary statistics
    of actual-token probabilities
    \citep{malinin2021uncertainty,ren2023outofdistribution}
    to learned detectors such as LOS-Net~\citep{HD-Guy}, which processes
    actual-token probabilities and ranked candidate distributions
    across the response. Its formulation can approximate a broad
    family of probability-based scoring functions.

    \paragraph{A blind spot in probability-only representations.}
    Despite this progression, probability-only gray-box detectors use probability values
    while discarding the identities of the tokens carrying them. Consequently,
    they reveal \emph{how uncertain} the model is, but not \emph{what it is
    uncertain about}. Two decoding states may have the same numerical probability
    profile while assigning that probability mass to entirely different tokens.
    A probability-only detector maps these states to the same representation,
    discarding potentially useful information about the alternatives considered
    by the model.
    
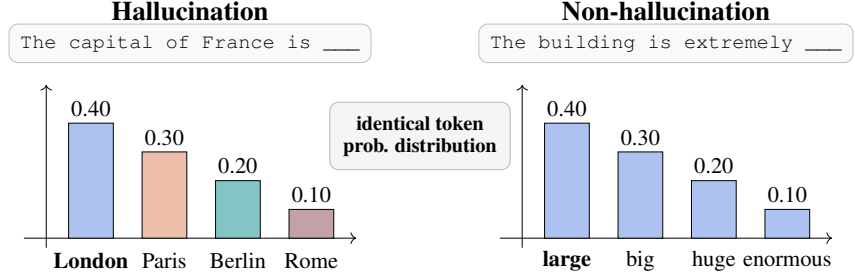
\begin{figure}[h]
    \centering
    \resizebox{0.8\linewidth}{!}{%
    \begin{tikzpicture}
    
    \node[font=\large\bfseries] at (2.25,3.55)
    {Hallucination};
    
    \node[
    draw=gray!50,
    rounded corners,
    fill=gray!5,
    font=\small,
    inner sep=5pt
    ] at (2.25,3.05)
    {\texttt{The capital of France is \_\_\_}};
    
    \draw[->] (-0.35,0) -- (4.85,0);
    \draw[->] (0,-0.15) -- (0,2.40);
    
    \draw[fill=myblue!45]
    (0.35,0) rectangle (1.05,1.80);
    \draw[fill=p1tds!45]
    (1.50,0) rectangle (2.20,1.35);
    \draw[fill=extteal!55]
    (2.65,0) rectangle (3.35,0.90);
    \draw[fill=embcolor!45]
    (3.80,0) rectangle (4.50,0.45);
    
    \node[font=\normalsize] at (0.70,2.03) {0.40};
    \node[font=\normalsize] at (1.85,1.58) {0.30};
    \node[font=\normalsize] at (3.00,1.13) {0.20};
    \node[font=\normalsize] at (4.15,0.68) {0.10};
    
    \node[font=\normalsize\bfseries] at (0.70,-0.35) {London};
    \node[font=\normalsize] at (1.85,-0.35) {Paris};
    \node[font=\normalsize] at (3.00,-0.35) {Berlin};
    \node[font=\normalsize] at (4.15,-0.35) {Rome};
    
    \node[
    draw=gray!55,
    rounded corners,
    fill=gray!8,
    align=center,
    font=\small\bfseries,
    inner sep=6pt
    ] at (5.85,1.6)
    {identical token\\prob. distribution};
    
    \begin{scope}[xshift=7.45cm]
    
    \node[font=\large\bfseries] at (2.25,3.55)
    {Non-hallucination};
    
    \node[
    draw=gray!50,
    rounded corners,
    fill=gray!5,
    font=\small,
    inner sep=5pt
    ] at (2.25,3.05)
    {\texttt{The building is extremely \_\_\_}};
    
    \draw[->] (-0.35,0) -- (4.85,0);
    \draw[->] (0,-0.15) -- (0,2.40);
    
    \draw[fill=myblue!45]
    (0.35,0) rectangle (1.05,1.80);
    \draw[fill=myblue!45]
    (1.50,0) rectangle (2.20,1.35);
    \draw[fill=myblue!45]
    (2.65,0) rectangle (3.35,0.90);
    \draw[fill=myblue!45]
    (3.80,0) rectangle (4.50,0.45);
    
    \node[font=\normalsize] at (0.70,2.03) {0.40};
    \node[font=\normalsize] at (1.85,1.58) {0.30};
    \node[font=\normalsize] at (3.00,1.13) {0.20};
    \node[font=\normalsize] at (4.15,0.68) {0.10};
    
    \node[font=\normalsize\bfseries] at (0.70,-0.35) {large};
    \node[font=\normalsize] at (1.85,-0.35) {big};
    \node[font=\normalsize] at (3.00,-0.35) {huge};
    \node[font=\normalsize] at (4.15,-0.35) {enormous};
    
    \end{scope}
    
    \end{tikzpicture}%
    }
    \caption{Identical numerical profiles can conceal different candidate structures. Both cases share the same token distribution, so a probability-only detector receives the same input. Candidate identities reveal competing factual answers on the left and paraphrastic alternatives on the right. The
    emitted token is shown in bold.}
    \label{fig:intro-identity}
    \end{figure}

    Figure~\ref{fig:intro-identity} illustrates this problem. Both decoding states
    have exactly the same ranked candidate token probabilities
    . They differ only in the
    tokens carrying those probabilities: one (left panel) contains competing factual answers,
    whereas the other (right panel) contains alternative, yet semantically equivalent, ways to complete the sentence -- potentially indicating a lower risk of hallucination. This
    distinction is visible only when candidate token identities are taken into account and is currently absent from all previous gray-box detectors that rely on
    probability-only representations.
    
    \paragraph{Our approach.}
    
    GURI is a lightweight gray-box detector that combines token
    identity codes with their probabilities and aggregates the resulting
    representations across the answer. The fusion and prediction
    components are trained jointly for answer-level hallucination
    detection.
    
\begin{figure*}[t]
    \centering
    \includegraphics[width=\textwidth]{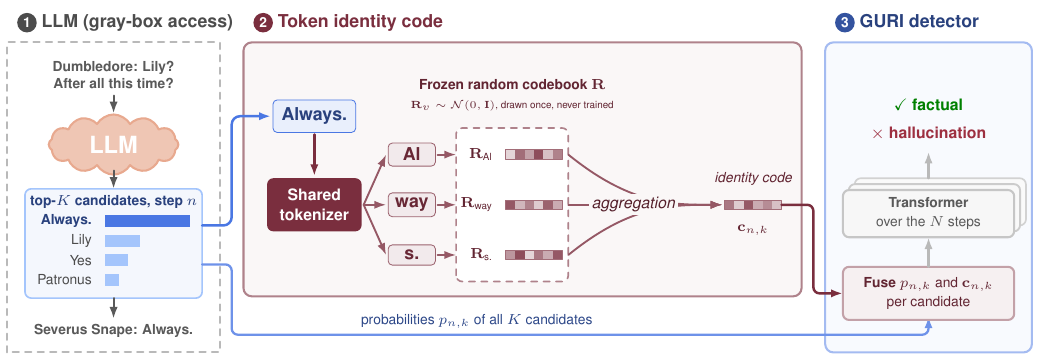}
    \caption{GURI pipeline. At each generation step, a shared external
tokenizer splits each top-$K$ candidate into subtokens whose codes
are aggregated. Fusion modules combine these identity codes with
candidate probabilities, and a response-level Transformer processes
the resulting sequence. Both components are trained jointly for
answer-level hallucination detection.}\label{fig:intro-pipeline}
    \end{figure*}

    Code vectors may be pretrained, learned, or fixed random.
    We use fixed random codes by default, as they match or outperform
    pretrained embeddings (Section~\ref{sec:representation-analysis}).
    To support cross-model transfer, we retokenize candidate strings
    with a shared external tokenizer and aggregate their subtoken
    vectors into identity codes. This common encoding also supports
    closed-source models exposing candidate strings and probabilities,
    without requiring their tokenizer or embedding table.
    See Section~\ref{sec:method} for encoding and architecture details
    and Figure~\ref{fig:intro-pipeline} for the pipeline.
    
    \paragraph{Results.}
    Across three open-weight models and three question-answering benchmarks,
    GURI outperforms both the Confidence baseline and LOS-Net in all nine
    model--dataset combinations. Mean test ROC-AUC increases from
    \(72.09\%\) for LOS-Net to \(79.09\%\) for GURI
    (Table~\ref{tab:main-results}).
    On GPT-4.1, mean ROC-AUC increases from \(67.39\%\) to \(71.84\%\),
    with an \(8.14\)-point gain on NQ-WC (Table~\ref{tab:gpt41}).
Among the three open-weight models, GURI outperforms transferred
LOS-Net in 17 of 18 directed cross-model transfers without target
fine-tuning. For the three open-weight models, cross-dataset
transfer between HotpotQA and TriviaQA with the model held fixed
improves mean ROC-AUC over transferred LOS-Net by 3.76 points
(Section~\ref{sec:transfer}).
Channel ablations with random codes further show that retaining
only non-actual candidate identities preserves the performance
of the full detector.
    
    
    \paragraph{Contributions.}
    Our contributions are threefold:
    \textbf{(1)} We identify token identity as a missing signal in probability-only
    gray-box detection.
    \textbf{(2)} We introduce GURI, a token-identity aware, transferable, gray-box detector applicable to both
    open- and closed-source models.
    \textbf{(3)} We demonstrate consistent empirical gains and show that identities provide predictive information beyond token probabilities.
    
    \section{Previous Work.}
    \paragraph{Definitions and scope.}
    The term ``hallucination'' has been used to describe several types of
    erroneous model behavior
    \citep{liu2021token,huang2023survey,ji2023survey,rawte2023troubling}.
    Surveys distinguish factual inaccuracies from failures of faithfulness,
    including contradictions and content unsupported by the provided
    context \citep{ji2023survey,huang2023survey}.
    Following the broad usage in error-detection research
    \citep{orgad2025llms}, we use the term to encompass factual errors,
    reasoning mistakes, and inconsistencies in model outputs.
    
    \paragraph{Hallucination detection.}
    One class of approaches consists of \textbf{black-box methods}, which
    operate on the model's prompts and generated text. SelfCheckGPT
    compares a response against additional sampled passages to identify
    unsupported statements \citep{manakul2023selfcheckgpt}.
    Semantic-uncertainty methods group sampled responses by meaning before
    measuring disagreement
    \citep{kuhn2023semantic,farquhar2024detecting}. Discrete semantic
    entropy estimates group probabilities from sampling frequencies,
    allowing text-only detection \citep{farquhar2024detecting}.
    \textbf{White-box methods} leverage internal model signals. These range
    from probing classifiers \citep{belinkov2022probing} trained on hidden
    representations to predict correctness
    \citep{azaria2023internal,orgad2025llms,snyder2024early}, to more complex
    architectures that process activation tensors across layers and tokens
    \citep{bar2026beyond}, or combine activations and attention patterns
    \citep{frasca2025neural}.
    INSIDE provides another approach, measuring the diversity of internal
    representations across sampled responses through their covariance
    spectrum \citep{chen2024inside}.

    Between these access settings, \textbf{Gray-box methods} use output-token probabilities without
hidden states or attention matrices. Earlier approaches include
length-normalized sequence likelihood for hallucination detection
in machine translation \citep{guerreiro2022looking} and
P(True), the probability of an affirmative response when the model
is asked to verify an answer \citep{kadavath2022language}.
Related pipelines trigger external validation at low confidence
\citep{varshney2023stitch} or combine and calibrate detection scores
\citep{valentin2024cost}. Semantics-aware methods include TokenSAR,
which weights generated-token log-probabilities by the semantic
effect of token removal \citep{duan-etal-2024-shifting}, and
Claim-Conditioned Probability (CCP), which combines alternative-word
probabilities with entailment and contradiction judgments to
estimate claim-level uncertainty \citep{fadeeva-etal-2024-fact}.
Learnable methods include lightweight classifiers using summary
statistics of evaluator-model probabilities
\citep{quevedo2024detecting}. LOS-Net \citep{HD-Guy} processes
sequences of actual-token probabilities and ranked candidate
distributions with a tailored architecture whose formulation can
approximate a broad family of prior probability-based scoring functions.
    
    GURI combines single-generation gray-box access with a learned detector
    over candidate-token identities and probabilities. Its fixed random
    identity codes retain information absent from probability-only representations, without requiring the explicit semantic similarity
    or entailment computations used by prior semantics-based method.
    
    \section{GURI: Identity-Aware Gray-Box Detection}
    \label{sec:method}
    
    This section presents GURI. We first fix notation for the signals a gray-box
    interface exposes, then show that existing detectors discard part of them, and
    finally describe how GURI recovers and uses the discarded part - token identities.
    
    \subsection{Setup, motivation and overview of the appraoch}
    
    \paragraph{Task and observable signals. }
    
    Given an input query \(q\) and a response
    \(\hat{g}=(w_1,\ldots,w_N)\) generated by a target language model, our goal is
    to predict an answer-level binary label indicating whether \(\hat{g}\) is
    factual or hallucinated. At generation step \(t\), the target model defines
    the next-token distribution
    \[
    p_t(\cdot)=p(\cdot\mid q,w_{<t}).
    \]
    Let \(c_{t,j}\) denote the candidate token at generation step \(t\)
    with probability rank \(j\), for \(j=1,\ldots,K\).
    The gray-box interface exposes the top-\(K\) candidate probabilities
    \(X^{K}_{t,j}\), where the superscript \(K\) indicates the number of
    retained candidates, and the probability \(a_t\) of the generated
    token \(w_t\):
    \begin{equation}
    \label{eq:graybox-inputs}
    X^{K}_{t,j}=p_t(c_{t,j}),
    \qquad
    a_t=p_t(w_t),
    \qquad
    X^{K}_{t,1}\geq\cdots\geq X^{K}_{t,K}.
    \end{equation}
    
    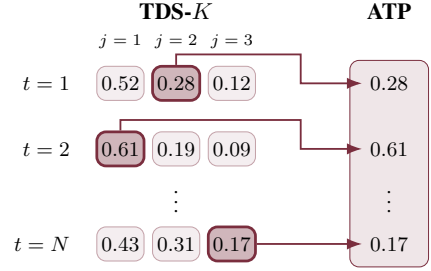
\begin{wrapfigure}{r}{0.4\textwidth}
    \centering
    \resizebox{\linewidth}{!}{%
    \begin{tikzpicture}[
      font=\small,
      box/.style={
        draw=embcolor!45,
        rounded corners,
        minimum width=0.72cm,
        minimum height=0.52cm,
        inner sep=1pt
      },
      other/.style={box,fill=embcolor!8},
      actual/.style={
        box,
        draw=embcolor,
        fill=embcolor!35,
        very thick
      },
      >=Latex
    ]
    \node[font=\bfseries] at (1.40,2.70) {TDS-$K$};
    \node[font=\bfseries] at (4.65,2.70) {ATP};
    
    \node[font=\scriptsize] at (0.55,2.25) {$j=1$};
    \node[font=\scriptsize] at (1.40,2.25) {$j=2$};
    \node[font=\scriptsize] at (2.25,2.25) {$j=3$};
    
    \node[left] at (-0.1,1.60) {$t=1$};
    \node[other] at (0.55,1.60) {$0.52$};
    \node[actual] (c1) at (1.40,1.60) {$0.28$};
    \node[other] at (2.25,1.60) {$0.12$};
    
    \node[left] at (-0.1,0.60) {$t=2$};
    \node[actual] (c2) at (0.55,0.60) {$0.61$};
    \node[other] at (1.40,0.60) {$0.19$};
    \node[other] at (2.25,0.60) {$0.09$};
    
    \node at (1.40,-0.10) {$\vdots$};
    
    \node[left] at (-0.1,-0.85) {$t=N$};
    \node[other] at (0.55,-0.85) {$0.43$};
    \node[other] at (1.40,-0.85) {$0.31$};
    \node[actual] (cN) at (2.25,-0.85) {$0.17$};
    
    \draw[draw=embcolor,fill=embcolor!12,rounded corners]
      (4.05,-1.20) rectangle (5.25,1.95);
    \node (a1) at (4.65,1.60) {$0.28$};
    \node (a2) at (4.65,0.60) {$0.61$};
    \node at (4.65,-0.10) {$\vdots$};
    \node (aN) at (4.65,-0.85) {$0.17$};
    
    \draw[->,embcolor,thick]
      (c1.north) -- ++(0,0.16)
      -| (3.30,1.60) -- (a1.west);
    \draw[->,embcolor,thick]
      (c2.north) -- ++(0,0.16)
      -| (3.30,0.60) -- (a2.west);
    \draw[->,embcolor,thick]
      (cN.east) -- (aN.west);
    \end{tikzpicture}%
    }
    \caption{TDS-$K$ and ATP inputs.}
    \label{fig:tds-atp}\end{wrapfigure}

    Across the \(N\) generation steps, these values form the top-\(K\) Token
    Distribution Sequence
    \(\mathbf{X}^{K}\in\mathbb{R}^{N\times K}\) and the Actual Token Probability
    sequence \(\mathbf{a}\in\mathbb{R}^{N}\), which we abbreviate as TDS-\(K\)
    and ATP, respectively. Together, \((\mathbf{X}^{K},\mathbf{a})\) constitute the top-\(K\)
    version of the LLM Output Signature (LOS) representation introduced
    We focus on this restricted representation because it reflects a realistic deployment setting, where the interface exposes only
    a limited number of candidate probabilities.
    Figure~\ref{fig:tds-atp} illustrates the distinction
    between TDS-\(K\) and ATP.
    
    \par
    Importantly, the gray-box interface exposes more than these numerical sequences: every
    probability is accompanied by the \emph{identity} of the token carrying it. We
    represent the complete observable generation trace as
    \begin{equation}
    \label{eq:complete-trace}
    \mathcal{O}
    =
    \left(
    \left\{(c_{t,j},X^{K}_{t,j})\right\}_{j=1}^{K},
    (w_t,a_t)
    \right)_{t=1}^{N}.
    \end{equation}

    \paragraph{Motivation for using token identities.} A main limitation in previous gray-box detectors, like \citep{HD-Guy} and its predecessors, is that they do not process \(\mathcal{O}\) directly.
    Instead, they operate on the projection
    \(\pi(\mathcal{O})=(\mathbf{X}^{K},\mathbf{a})\),
    which retains the probability values while discarding the token identities
    and their assignment to those values.
    This projection is not injective. Two distinct traces may satisfy \(
    \mathcal{O}\neq\mathcal{O}',
    \pi(\mathcal{O})=\pi(\mathcal{O}'),
    \)
    in which case every probability-only detector \(h\) must satisfy
    \(
    h\!\left(\pi(\mathcal{O})\right)
    =
    h\!\left(\pi(\mathcal{O}')\right).
    \)
    The resulting ambiguity, illustrated in Figure~\ref{fig:intro-identity},
    is a property of the representation rather than of any particular detector: no (deterministic) downstream architecture can recover distinctions
    that its input has removed.
    
    GURI encodes token strings, fuses their codes with probabilities,
    and aggregates the resulting sequence to predict answer factuality
    (Figure~\ref{fig:intro-pipeline}).
    
    
    \subsection{Token-Identity Codes}
    To combine token identities with probabilities, we retokenize
each token string using a shared external tokenizer and aggregate
its subtoken codes into one vector. This ensures that the same
string receives the same representation across generation steps,
answers, and target models.
    \paragraph{From candidate strings to codes via an external tokenizer.}
    Alongside each exposed probability, we retain the string of the corresponding
    token. We denote the candidate strings by
    \(
    s_{t,j}=\operatorname{text}(c_{t,j}),
    \)
    the $j$ ranked candidate token string. And the actual-token string by
    \(
    s^{\mathrm{w}}_t=\operatorname{text}(w_t).
    \)
    These strings can originate from models with different native vocabularies and
    tokenizers. GURI therefore maps every observed string through a shared
    external tokenizer.
    
    Formally, let
    \(
    \mathcal{T}:\Sigma^*\rightarrow V_{\mathrm{ext}}^*
    \)
    map an arbitrary string \(s\) to a sequence of external (sub)tokens
    \(\mathcal{T}(s)=(v_1,\ldots,v_m)\). Given a fixed code table
    \(C:V_{\mathrm{ext}}\rightarrow\mathbb{R}^{d}\), our default random-code configuration represents the original
    string by mean-pooling its subtoken codes and
    \(\ell_2\)-normalizing the result:
    \begin{equation}
    \label{eq:identity-code}
    \begin{aligned}
    \bar e_C(s)
    &=
    \frac{1}{m}\sum_{i=1}^{m}C(v_i),\qquad
    e_C(s)
    =
    \frac{\bar e_C(s)}
    {\left\|\bar e_C(s)\right\|_2},
    \end{aligned}
    \qquad m>0.
    \end{equation}
    This construction maps tokens from different target vocabularies into a common
    representation space without requiring any alignment between their native
    token IDs. The shared encoding \(e_C\) assigns the same representation to
    every occurrence of a given token string. Other realizations of \(e_C\) are possible, for instance, replacing the
    mean-pooling step with a small learned network over the subtoken codes. We used this construction for the default random codes.
    For pretrained embeddings, we used the unnormalized subtoken
    mean (Appendix~\ref{app:identity-code-tables}).
    
    \paragraph{Instantiations of code tables.}
    The construction in Equation~\ref{eq:identity-code} requires only a fixed
    vector for every external subtoken. Our primary GURI configuration uses a
    random table sampled once and then frozen. The resulting vectors provide
    identifiers but contain no pretrained geometric or semantic relationships between
    tokens. Pretrained embeddings provide another possible choice of code table,
    but do not outperform fixed random codes in our experiments
    (Section~\ref{sec:representation-analysis}).
    
    \subsection{Architecture}We now describe the architecture that processes the probabilities and the
    identity codes together. In short, GURI combines each probability with the
    identity code of the token carrying it, and aggregates the fused sequence across
    the generated answer using a transformer.
    \paragraph{Creating ATP and TDS aligned identity tensors.}
    The first step is to arrange the identity codes so that each one sits alongside
    the probability it belongs to.
    Applying \(e_C\) to every observed token string yields the candidate identity
    tensor and actual-token identity sequence
    \begin{equation}
    \mathbf{E}^{K}_{t,j}=e_C(s_{t,j}),
    \qquad
    \mathbf{E}^{\mathrm{w}}_t=e_C(s^{\mathrm{w}}_t),
    \end{equation}
    where
    \(\mathbf{E}^{K}\in\mathbb{R}^{N\times K\times d}\) is aligned entry by
    entry with the $k$-TDS \(\mathbf{X}^{K}\), and
    \(\mathbf{E}^{\mathrm{w}}\in\mathbb{R}^{N\times d}\) is aligned with the ATP
    sequence \(\mathbf{a}\).
    
    \paragraph{Fusing probability and identity information.}
    In the next step, GURI combines each probability with the identity code of the token carrying
    it. Separate shared MLPs process the candidate and actual-token channels:\begin{align}
    \widetilde{X}_{t,j}
    =
    f_{\mathrm{cand}}
    \left(
    [X^{K}_{t,j};\mathbf{E}^{K}_{t,j}]
    \right),\qquad
    \widetilde{a}_t
    &=
    f_{\mathrm{act}}
    \left(
    [a_t;\mathbf{E}^{\mathrm{w}}_t]
    \right),
    \end{align}
    where \([\,\cdot\,;\,\cdot\,]\) denotes concatenation and
    \(
    f_{\mathrm{cand}},f_{\mathrm{act}}
    :
    \mathbb{R}^{d+1}\rightarrow\mathbb{R}.
    \)
    The same candidate MLP is applied to every rank and generation step, and the
    same actual-token MLP is applied throughout the response.
   Scalar fusion preserves the probability-only input shapes.
We also evaluate 256-dimensional token representations using
an alternating Transformer (Section~\ref{sec:alternating-ablation}).

    \paragraph{GURI's backbone architecture.}
Our response-level encoder follows the architecture of
LOS-Net \citep{HD-Guy}, which we briefly describe here
for completeness.
A Transformer \(H_\theta\) aggregates
\(\widetilde{\mathbf{X}}\) and \(\widetilde{\mathbf{a}}\)
across the \(N\) generation steps to predict answer factuality:    \begin{equation}
    \label{eq:guri-prediction}
    \widehat{y}
    =
    H_\theta
    \left(
    \widetilde{\mathbf{X}},
    \widetilde{\mathbf{a}}
    \right).
    \end{equation}
    Here, \(\widehat{y}\) is the predicted probability that the complete response
    is factual. The candidate-fusion MLP, actual-token-fusion MLP, and
    Transformer are optimized jointly under the answer-level training objective;
    only the external code table remains fixed.

    
    
    \section{Experiments}
    \label{sec:experiments}
    
    \subsection{Experimental Setup}
    
    \paragraph{Models, tasks, and labels.}
    We evaluate three open-weight target models:
    Gemma-4-12B-IT, Llama-3.2-3B-Instruct, and
    Ministral-3-8B-Instruct. Our evaluation uses three question-answering
    benchmarks---HotpotQA, TriviaQA, and Natural Questions with context
    (NQ-WC)---for a total of nine model--dataset combinations. HotpotQA and
    TriviaQA are evaluated in a closed-book setting without supporting passages,
    whereas NQ-WC supplies the model with the associated context.
    
    Unless
    otherwise stated, all detectors use the top \(K=10\) candidates at each
    generation step.
    
    
    \paragraph{Main GURI configuration.}
    Unless otherwise stated, GURI uses the Phi-4-mini tokenizer as its shared
    external tokenizer and assigns a fixed random code to every subtoken in its
    vocabulary. Each candidate string is retokenized with this tokenizer, and the
    resulting subtoken codes are mean-pooled and \(\ell_2\)-normalized as in
    Equation~\ref{eq:identity-code}. The same tokenizer and random code table are
    used for every target model, allowing tokens from different native
    vocabularies to be represented in a common space. The code table remains
    frozen throughout training; only the probability--identity fusion networks and
    the sequence encoder are optimized.
    
    
    \paragraph{Baselines.}
    We compare GURI with two probability-only baselines. \textbf{Confidence}
    represents classical confidence scoring: for each model--dataset combination,
    we select on the validation set the strongest of the mean, minimum, and maximum
    of the ATP probabilities and log-probabilities. \textbf{LOS-Net} is the
    strongest learned probability-only gray-box baseline. It processes the
    TDS-\(K\) and ATP sequences across the generated answer but receives no token
    identities.
    
    \paragraph{Evaluation protocol.}
    Our primary metric is test ROC-AUC, reported in percent. For the main comparison, we report the mean and standard deviation
    of test ROC-AUC across all 36 hyperparameter--seed combinations
    in the grid (Appendix~\ref{app:model-selection}). Checkpoints are
    selected using ROC-AUC on the held-out validation set. Confidence baselines are
    deterministic once the validation-selected statistic has been chosen.
    Permutation experiments keep the trained detector fixed. Channel-removal experiments retrain it.
    
    
    \subsection{Main Results}
    \label{sec:main-results}
    
    \paragraph{Open-weight models.}
    
    Table~\ref{tab:main-results} compares the default random-code GURI with the
    probability-only baselines. GURI improves over LOS-Net in all nine displayed
    model--dataset combinations, with an average gain of approximately \(7.00\)
    ROC-AUC points. The improvements range from \(2.11\) points on
    Ministral/TriviaQA to \(12.30\) points on Gemma/NQ-WC.
    
    \begin{table*}[t]
    \caption{Hallucination-detection performance on the \(3\times3\)
    evaluation suite. Values are test ROC-AUC (\%). Learned detectors report the mean and standard deviation
    over all 36 hyperparameter--seed combinations. GURI uses a fixed
    random code table.}
    \label{tab:main-results}
    \centering
    \setlength{\tabcolsep}{8pt}
    \begin{tabular}{llccc}
    \toprule
    Dataset & Target model & Confidence & LOS-Net & GURI \\
    \midrule
    \multirow{3}{*}{HotpotQA}
    & Ministral & 64.98 & \(70.68_{\pm0.97}\) & \(\mathbf{77.18}_{\pm1.37}\) \\
    & Gemma     & 61.13 & \(74.11_{\pm1.30}\) & \(\mathbf{83.93}_{\pm0.66}\) \\
    & Llama     & 60.94 & \(80.48_{\pm1.01}\) & \(\mathbf{86.36}_{\pm0.58}\) \\
    \midrule
    \multirow{3}{*}{NQ-WC}
    & Ministral & 59.28 & \(65.01_{\pm1.00}\) & \(\mathbf{73.32}_{\pm0.84}\) \\
    & Gemma     & 55.29 & \(65.88_{\pm1.19}\) & \(\mathbf{78.18}_{\pm0.56}\) \\
    & Llama     & 53.27 & \(63.36_{\pm0.58}\) & \(\mathbf{73.06}_{\pm0.39}\) \\
    \midrule
    \multirow{3}{*}{TriviaQA}
    & Ministral & 64.40 & \(73.91_{\pm0.79}\) & \(\mathbf{76.02}_{\pm0.72}\) \\
    & Gemma     & 67.29 & \(77.75_{\pm0.49}\) & \(\mathbf{81.80}_{\pm0.66}\) \\
    & Llama     & 55.08 & \(77.63_{\pm0.81}\) & \(\mathbf{81.99}_{\pm0.15}\) \\
    \bottomrule
    \end{tabular}
    \end{table*}
    These results show that fixed token codes provide predictive information that
    is systematically absent from the numerical probability profile.
    
    \paragraph{Closed-source model.}
    
    We additionally evaluate GURI on GPT-4.1, for which neither hidden states nor
    the target embedding table are available. In the currently available
    experiments, GURI increases mean ROC-AUC from \(67.39\) to \(71.84\).
    The improvement is substantial on HotpotQA and NQ-WC, while performance on
    TriviaQA is approximately unchanged.
    
    \begin{table}[H]
    \caption{Test ROC-AUC (\%) on the closed-source GPT-4.1 model.}
    \label{tab:gpt41}
    \centering
    \begin{tabular}{lccc}\toprule
    Dataset & LOS-Net & GURI & Difference \\
    \midrule
    HotpotQA & \(64.20_{\pm0.43}\) & \(\mathbf{70.12}_{\pm0.36}\) & \(+5.92\) \\
    TriviaQA & \(\mathbf{72.49}_{\pm0.39}\) & \(71.78 _{\pm0.74}\) & \(-0.71\) \\
    NQ-WC    & \(65.49_{\pm0.23}\) & \(\mathbf{73.63}_{\pm0.61}\) & \(+8.14\) \\
    \midrule
    
    Mean     & 67.39 & \(\mathbf{71.84}\) & \(+4.45\) \\
    \bottomrule
    \end{tabular}
    \end{table}

    
    
    

    \subsection{Transfer Across Models and Datasets}\label{sec:transfer}

    \paragraph{Cross-model transfer.}
    We evaluated cross-model transfer within each dataset, applying
    a detector trained on one model's responses to another model's
    test responses without fine-tuning (zero-shot). GURI used the shared unit-norm
    random codes.

\begin{figure}[h]
\centering
\begingroup
\definecolor{guriHeatPositive}{RGB}{217,130,88}
\definecolor{guriHeatNegative}{RGB}{115,145,201}

\newcommand{\guriTransferPanel}[3]{%
\begin{scope}[xshift=#1cm]
\node[font=\bfseries\large] at (1.425,3.25) {#2};
\foreach \r/\c/\auc/\gain/\marked in {#3}{
    \pgfmathsetmacro{\cellX}{0.95*\c}
    \pgfmathsetmacro{\cellY}{0.95*(2-\r)}
    \pgfmathsetmacro{\cellShade}{min(100,abs(\gain)/30*100)}
    \ifdim \gain pt<0pt
        \fill[guriHeatNegative!\cellShade!white]
            (\cellX,\cellY) rectangle ++(0.95,0.95);
    \else
        \fill[guriHeatPositive!\cellShade!white]
            (\cellX,\cellY) rectangle ++(0.95,0.95);
    \fi
    \draw[white,line width=1pt]
        (\cellX,\cellY) rectangle ++(0.95,0.95);
    \node[font=\scriptsize\bfseries]
        at ({\cellX+0.475},{\cellY+0.56})
        {\pgfmathprintnumber[fixed,precision=2,zerofill]{\auc}};
    \node[font=\tiny]
        at ({\cellX+0.475},{\cellY+0.20})
        {\pgfmathprintnumber[fixed,precision=1,zerofill,showpos]{\gain}};
    \ifnum\marked=1
        \draw[black,line width=0.7pt]
            ({\cellX+0.035},{\cellY+0.035})
            rectangle ++(0.88,0.88);
        \node[font=\scriptsize,inner sep=0pt]
            at ({\cellX+0.78},{\cellY+0.78}) {$\bigstar$};
    \fi
}
\draw[black,line width=0.6pt] (0,0) rectangle (2.85,2.85);
\foreach \i/\model in {0/G,1/L,2/M}{
    \node[anchor=east,font=\scriptsize]
        at (-0.15,{2.375-0.95*\i}) {\model};
    \node[anchor=north,font=\scriptsize]
        at ({0.475+0.95*\i},-0.13) {\model};
}
\node[font=\small\bfseries] at (1.425,-0.95) {Target};
\end{scope}
}

\resizebox{\linewidth}{!}{%
\begin{tikzpicture}
\node[rotate=90,font=\small\bfseries] at (-1.30,1.80) {Source};

\guriTransferPanel{0}{HotpotQA}{
    0/0/84.84/23.71/1,
    0/1/64.43/3.49/0,
    0/2/74.98/10.00/1,
    1/0/74.88/13.75/0,
    1/1/87.31/26.37/1,
    1/2/68.87/3.89/0,
    2/0/73.99/12.86/0,
    2/1/49.74/-11.20/0,
    2/2/78.77/13.79/1
}

\guriTransferPanel{4.0}{TriviaQA}{
    0/0/82.69/15.40/1,
    0/1/69.75/14.67/0,
    0/2/69.30/4.90/0,
    1/0/73.37/6.08/0,
    1/1/81.97/26.89/1,
    1/2/68.26/3.86/0,
    2/0/76.53/9.24/0,
    2/1/73.12/18.04/0,
    2/2/77.56/13.16/1
}

\guriTransferPanel{8.0}{NQ-WC}{
    0/0/79.00/23.71/1,
    0/1/70.41/17.14/1,
    0/2/67.85/8.57/1,
    1/0/67.41/12.12/0,
    1/1/73.33/20.06/1,
    1/2/65.62/6.34/0,
    2/0/69.28/13.99/1,
    2/1/68.26/14.99/1,
    2/2/73.96/14.68/1
}

\foreach \i in {0,...,60}{%
    \pgfmathtruncatemacro{\legendGain}{\i-30}
    \pgfmathsetmacro{\legendTint}{abs(\legendGain)/30*100}
    \pgfmathsetmacro{\legendY}{0.85+2.0*\i/61}
    \ifdim\legendGain pt<0pt
        \fill[guriHeatNegative!\legendTint!white]
            (11.25,\legendY) rectangle ++(0.23,{2.0/61+0.002});
    \else
        \fill[guriHeatPositive!\legendTint!white]
            (11.25,\legendY) rectangle ++(0.23,{2.0/61+0.002});
    \fi
}
\draw[line width=0.5pt] (11.25,0.85) rectangle (11.48,2.85);
\foreach \value/\height in {
    -30/0.85,-15/1.35,0/1.85,+15/2.35,+30/2.85
}{
    \draw[line width=0.4pt]
        (11.48,\height) -- (11.55,\height);
    \node[anchor=west,font=\scriptsize]
        at (11.59,\height) {\value};
}
\node[rotate=90,font=\scriptsize]
    at (12.40,1.85) {Gain over Confidence (AUC points)};
\node[font=\small] at (11.35,-0.30) {$\bigstar$};
\node[anchor=west,align=left,font=\scriptsize]
    at (11.65,-0.30) {Above target-trained\\LOS-Net};
\end{tikzpicture}%
}
\endgroup
\caption{Cross-model transfer of GURI with unit-norm random codes.
Rows and columns denote source and target models:
G, L, and M denote Gemma, Llama, and Ministral.
Each row uses one source-validation-selected checkpoint.
Bold values show test ROC-AUC (\%). Colors and signed values show
gains over target-model Confidence (AUC points).
Stars and outlines mark scores above target-trained LOS-Net,
not statistical significance.}
\label{fig:cross-model-transfer}
\end{figure}
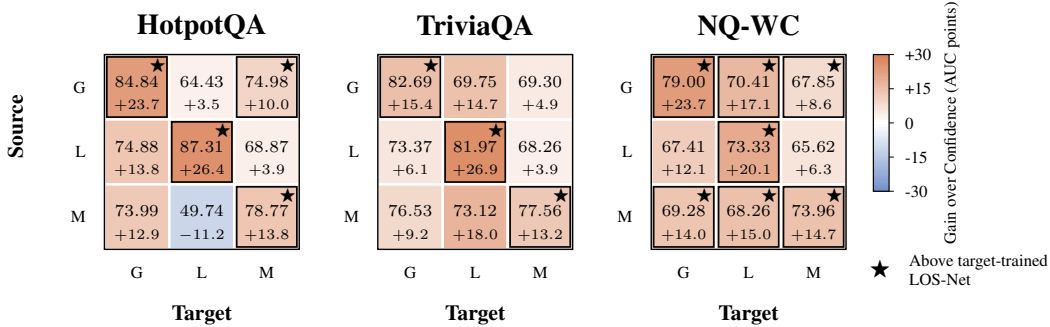

    Figure~\ref{fig:cross-model-transfer} summarizes GURI's cross-model transfer results.
    Complete numerical results for both GURI and LOS-Net are reported
    in Figure~\ref{tab:cross-model-transfer} in
    Appendix~\ref{app:cross-model-transfer}, alongside the evaluation protocol.
    

    Among the three open-weight models, GURI outperformed transferred
LOS-Net in 17 of 18 directed cross-model transfers. Its mean advantage
was 13.52 ROC-AUC points on NQ-WC, 11.93 on HotpotQA, and 11.27 on
TriviaQA. It exceeded target-model Confidence in 17 of 18 transfers
(LOS-Net: 7 of 18) and target-trained LOS-Net in five.

    \paragraph{Cross-dataset and joint transfer.}
    We also evaluated transfer between datasets, either keeping the
    target model fixed or changing both model and dataset.
    Between HotpotQA and TriviaQA, GURI improved over LOS-Net
by 3.76 ROC-AUC points on average with the model fixed,
and by 4.94 points when the model also changed.
For transfers involving NQ-WC, GURI achieved mean ROC-AUCs of
\(52.94\%\) and \(50.99\%\), respectively, compared with
\(52.97\%\) and \(51.56\%\) for LOS-Net.
    Joint model--dataset transfer did not exceed the target-model
    Confidence baseline on average in either dataset group.
    Since NQ-WC supplied context while HotpotQA and TriviaQA did not,
    these comparisons do not isolate dataset changes from changes
    in context availability.
    Table~\ref{tab:dataset-transfer-summary} summarizes these results.
    Appendix~\ref{app:cross-dataset-transfer} provides the full matrices
    and protocol.
    
    \subsection{Analysis and Ablations}
    \label{sec:analysis}
    
    \subsubsection{Does the Source of the Codes Matter? }
    \label{sec:representation-analysis}
    
    We initially hypothesized that pretrained embeddings would improve
detection by providing semantic relationships between tokens.
We compare fixed random codes with pretrained Phi-4-mini embeddings
using the same external tokenizer, mean-pooling, detector architecture,
and training protocol. Random codes are unit-normalized; Phi embeddings
are not (Appendix~\ref{app:identity-code-tables}).
Section~\ref{sec:alternating-ablation} additionally examines
vector-valued representations using an alternating Transformer.

\begin{table}[H]
\caption{Test ROC-AUC (\%) for code-source and architecture variants.
Random codes were unit-normalized; Phi-4-mini embeddings were not.
Bold indicates the highest mean in each row, including ties
at the reported precision.}
\label{tab:code-source}
\label{tab:alternating-ablation}
\centering
\small
\setlength{\tabcolsep}{4pt}
\begin{tabular}{llcccc}
\toprule
Dataset & Target model & GURI: Random & GURI: Phi
& Alternating: Phi & Alternating: Random \\
\midrule
\multirow{3}{*}{HotpotQA}
& Ministral
& $\mathbf{77.18}_{\pm1.37}$ & $76.82_{\pm0.78}$
& $\mathbf{77.18}_{\pm1.35}$ & $76.85_{\pm1.24}$ \\
& Gemma
& $83.93_{\pm0.66}$ & $81.70_{\pm0.60}$
& $82.91_{\pm0.50}$ & $\mathbf{84.61}_{\pm0.14}$ \\
& Llama
& $86.36_{\pm0.58}$ & $85.60_{\pm0.56}$
& $86.19_{\pm0.18}$ & $\mathbf{87.55}_{\pm0.10}$ \\
\midrule
\multirow{3}{*}{NQ-WC}
& Ministral
& $\mathbf{73.32}_{\pm0.84}$ & $70.53_{\pm1.04}$
& $71.14_{\pm0.40}$ & $72.94_{\pm0.64}$ \\
& Gemma
& $\mathbf{78.18}_{\pm0.56}$ & $76.58_{\pm0.67}$
& $76.72_{\pm0.84}$ & $76.99_{\pm1.25}$ \\
& Llama
& $73.06_{\pm0.39}$ & $69.87_{\pm0.85}$
& $69.62_{\pm0.51}$ & $\mathbf{73.43}_{\pm0.71}$ \\
\midrule
\multirow{3}{*}{TriviaQA}
& Ministral
& $76.02_{\pm0.72}$ & $75.84_{\pm1.00}$
& $\mathbf{76.84}_{\pm0.84}$ & $75.13_{\pm1.06}$ \\
& Gemma
& $81.80_{\pm0.66}$ & $80.90_{\pm0.34}$
& $81.01_{\pm0.49}$ & $\mathbf{83.30}_{\pm0.12}$ \\
& Llama
& $81.99_{\pm0.15}$ & $81.10_{\pm0.44}$
& $80.86_{\pm0.33}$ & $\mathbf{82.04}_{\pm0.21}$ \\
\bottomrule
\end{tabular}
\end{table}

    With scalar fusion, random codes achieve higher mean ROC-AUC
than pretrained Phi embeddings in all nine settings
(Table~\ref{tab:code-source}). Pretrained embeddings are therefore
not required for the gains observed with this architecture.
    

    \subsubsection{Does Early Scalar Compression Limit Gains?}
    \label{sec:alternating-ablation}
    Inspired by ACT-ViT's processing of structured activation tensors
    \citep{bar2026beyond}, we evaluate an alternating Transformer with
    fixed random codes and Phi embeddings to examine whether early scalar compression
    limits their usefulness.

  We view this variant as a natural extension of probability-only
sequence models to vector-valued candidate representations, with
attention alternating across generation positions and candidate slots.
    Architectural details are provided in
    Appendix~\ref{app:alternating-transformer}.

    We compare each alternating variant with scalar-fusion GURI
using the corresponding code source.

Retaining 256-dimensional token representations does not yield a
consistent improvement over scalar fusion across code sources.
Random codes also outperform Phi embeddings in seven of nine
alternating settings, providing no evidence that scalar fusion
masks a benefit of pretrained embeddings.

    

    \subsubsection{The Role of Actual and Non-Actual Token Identities}
    \label{sec:channel-retraining}
    
    
   Actual-token identities describe the generated sequence;
non-actual identities describe its alternatives. We retrain
GURI with either channel alone or both. All probability inputs remain unchanged, and removed identity codes are set to zero.
    Evaluation-time permutation experiments are reported in
    Appendix~\ref{sec:permutation-analysis}.
    \begin{figure*}[h]
    \centering
    \begin{tikzpicture}
    
    \begin{groupplot}[
        group style={
            group size=3 by 1,
            horizontal sep=1.1cm,
            ylabels at=edge left
        },
        width=0.335\textwidth,
        height=5.3cm,
        xmin=0.6,
        xmax=3.4,
        xtick={1,2,3},
        xticklabels={Gemma,Llama,Ministral},
        tick label style={font=\scriptsize},
        title style={font=\small\bfseries},
        ylabel style={font=\small},
        ymajorgrids=true,
        grid style={gray!20},
        axis line style={gray!70},
        legend style={
            font=\scriptsize,
            draw=none,
            legend columns=4,
            column sep=7pt
        }
    ]
    
    \nextgroupplot[
        title={TriviaQA},
        ymin=72,
        ymax=84,
        ytick={72,74,76,78,80,82,84},
        ylabel={Test ROC-AUC},
        legend to name=channelretraininglegend
    ]
    
    \addplot+[
        only marks,
        mark=*,
        mark size=2.5pt,
        color=gray!70!black,
        error bars/.cd,
        y dir=both,
        y explicit
    ] coordinates {
        (0.82,78.2) +- (0,0.6)
        (1.82,77.9) +- (0,0.5)
        (2.82,74.2) +- (0,0.7)
    };
    \addlegendentry{LOS-Net}
    
    \addplot+[
        only marks,
        mark=square*,
        mark size=2.5pt,
        color=myblue,
        error bars/.cd,
        y dir=both,
        y explicit
    ] coordinates {
        (0.94,79.9) +- (0,0.1)
        (1.94,81.2) +- (0,0.0)
        (2.94,75.4) +- (0,0.1)
    };
    \addlegendentry{Actual only}
    
    \addplot+[
        only marks,
        mark=triangle*,
        mark size=3pt,
        color=extteal,
        error bars/.cd,
        y dir=both,
        y explicit
    ] coordinates {
        (1.06,83.0) +- (0,0.1)
        (2.06,82.6) +- (0,0.0)
        (3.06,78.3) +- (0,0.4)
    };
    \addlegendentry{Non-actual only}
    
    \addplot+[
        only marks,
        mark=diamond*,
        mark size=3pt,
        color=embcolor,
        error bars/.cd,
        y dir=both,
        y explicit
    ] coordinates {
        (1.18,82.1) +- (0,0.1)
        (2.18,82.0) +- (0,0.2)
        (3.18,75.9) +- (0,0.5)
    };
    \addlegendentry{Both}
    
    \nextgroupplot[
        title={HotpotQA},
        ymin=68,
        ymax=90,
        ytick={68,72,76,80,84,88}
    ]
    
    \addplot+[
        only marks,
        mark=*,
        mark size=2.5pt,
        color=gray!70!black,
        error bars/.cd,
        y dir=both,
        y explicit
    ] coordinates {
        (0.82,75.7) +- (0,0.6)
        (1.82,80.7) +- (0,0.0)
        (2.82,71.4) +- (0,0.7)
    };
    
    \addplot+[
        only marks,
        mark=square*,
        mark size=2.5pt,
        color=myblue,
        error bars/.cd,
        y dir=both,
        y explicit
    ] coordinates {
        (0.94,82.3) +- (0,0.1)
        (1.94,84.8) +- (0,0.1)
        (2.94,75.3) +- (0,0.3)
    };
    
    \addplot+[
        only marks,
        mark=triangle*,
        mark size=3pt,
        color=extteal,
        error bars/.cd,
        y dir=both,
        y explicit
    ] coordinates {
        (1.06,84.3) +- (0,0.0)
        (2.06,87.5) +- (0,0.1)
        (3.06,78.7) +- (0,0.5)
    };
    
    \addplot+[
        only marks,
        mark=diamond*,
        mark size=3pt,
        color=embcolor,
        error bars/.cd,
        y dir=both,
        y explicit
    ] coordinates {
        (1.18,84.1) +- (0,0.5)
        (2.18,86.1) +- (0,0.5)
        (3.18,77.2) +- (0,1.5)
    };
    
    \nextgroupplot[
        title={NQ (with context)},
        ymin=60,
        ymax=82,
        ytick={60,64,68,72,76,80}
    ]
    
    \addplot+[
        only marks,
        mark=*,
        mark size=2.5pt,
        color=gray!70!black,
        error bars/.cd,
        y dir=both,
        y explicit
    ] coordinates {
        (0.82,65.5) +- (0,1.4)
        (1.82,63.4) +- (0,0.1)
        (2.82,64.7) +- (0,2.2)
    };
    
    \addplot+[
        only marks,
        mark=square*,
        mark size=2.5pt,
        color=myblue,
        error bars/.cd,
        y dir=both,
        y explicit
    ] coordinates {
        (0.94,75.5) +- (0,0.2)
        (1.94,72.1) +- (0,0.0)
        (2.94,72.4) +- (0,0.2)
    };
    
    \addplot+[
        only marks,
        mark=triangle*,
        mark size=3pt,
        color=extteal,
        error bars/.cd,
        y dir=both,
        y explicit
    ] coordinates {
        (1.06,79.3) +- (0,0.2)
        (2.06,74.3) +- (0,0.1)
        (3.06,74.7) +- (0,0.7)
    };
    
    \addplot+[
        only marks,
        mark=diamond*,
        mark size=3pt,
        color=embcolor,
        error bars/.cd,
        y dir=both,
        y explicit
    ] coordinates {
        (1.18,78.5) +- (0,0.4)
        (2.18,73.1) +- (0,0.3)
        (3.18,73.6) +- (0,0.5)
    };
    
    \end{groupplot}
    
    \node[anchor=north] at
    ($(group c1r1.south)!0.5!(group c3r1.south)+(0,-0.8cm)$)
    {\ref{channelretraininglegend}};
    
    \end{tikzpicture}
    
    \caption{
    \textbf{Non-actual identities alone match or exceed the full model
    in all nine settings.}
    Test ROC-AUC after retraining with unit-norm random codes and
    \emph{actual only}, \emph{non-actual only}, or \emph{both} identity
    channels, retaining all probability inputs.
    Markers and error bars show the mean and standard deviation
    over three training seeds. LOS-Net was retrained with the same
    hyperparameters. Actual identities alone recover only part of the gain.
    }\label{fig:channel-retraining}
    \end{figure*}
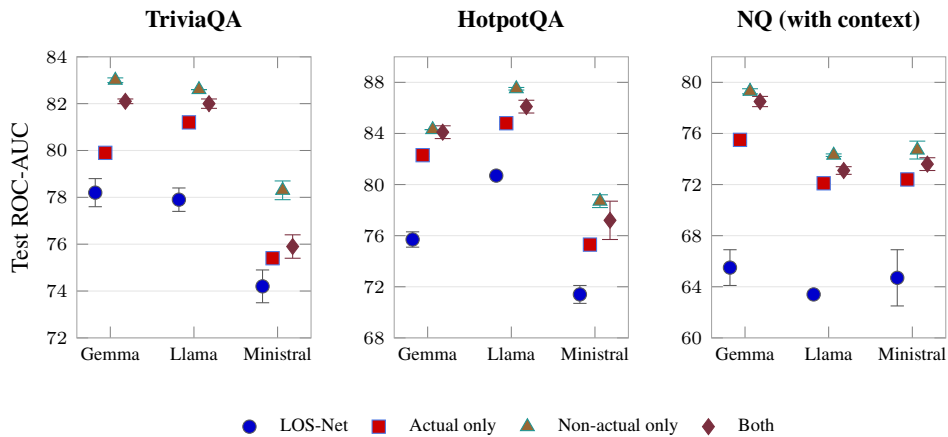

    Figure~\ref{fig:channel-retraining} shows the outcome. In all nine model--dataset combinations the ordering is the same:
    \emph{non-actual only} $\geq$ \emph{both} $>$ \emph{actual only} $>$ LOS-Net.
    A detector that never sees the identity of the actual token matches or
    slightly exceeds the full model, whereas a detector that sees only the actual
    token stays 0.5 to 3 points below it.
    
    Repeating the experiment with pretrained Phi-4-mini embeddings
    (Fig.~\ref{fig:channel-retraining-phi}) changes the picture in one specific place:
    the \emph{non-actual only} detector drops by 2 to 7 points in every cell,
    while the \emph{actual only} detector moves by at most 3. With pretrained
    codes the two channels become complementary, and on NQ-WC the actual token
    alone is worth more than the non-actual candidates; the full model
    nevertheless stays below its random-code counterpart in every cell.

    \section{Discussion and Limitations}
    \label{sec:discussion}
    
    Our results support token identity as a useful predictive signal beyond
    next-token probabilities. 
    
    
    The random-code results also require careful interpretation. The main random
    construction retains the compositional structure induced by the external
    tokenizer: target tokens that share external subtokens also share code
    components. Accordingly, the experiments establish that pretrained semantic geometry is
    unnecessary for the main gains, not that every form of lexical or tokenization
    structure is irrelevant.
    
    The channel analyses cover the nine model--dataset combinations of our suite
    under a single training configuration, and the ordering between the two
    identity channels depends on the code table: with random codes the non-actual
    candidates carry the larger value, with pretrained embeddings the two channels
    are complementary. The ordering should therefore not be read as a universal
    property of the channels. Evaluation-time permutations additionally create
    shifted inputs and measure reliance, whereas retraining measures what a
    channel can learn when available from the beginning; the two interventions
    answer different questions.
    
    Finally, the evaluation relies on automatic string-match labels, which may
    reward a correct substring inside an otherwise incorrect response and penalize
    appropriate abstention. 
    In the open-weight experiments, detector inputs and labeling
    included continuation after end tokens; the effect of stopping
    at the first end token remains unevaluated
    (Appendix~\ref{app:prompts-generation}).
    The method also assumes access to candidate token
    strings and probabilities, which are not exposed by every commercial API.
    Future work should evaluate identity-aware detection under human factuality
    labels, additional generation protocols, and broader model and task families.

    \section{Conclusion}
    \label{sec:conclusion}
    
GURI augments probability-only gray-box detection with fixed
token-identity codes while preserving its access and efficiency
advantages. Across our evaluation suite, it consistently improves
over LOS-Net. Fixed random codes perform comparably to pretrained
embeddings, showing that the gains do not require pretrained
semantic geometry. Channel ablations show that non-actual candidate
identities alone preserve full performance with random codes.
These results demonstrate the value of token identities for
hallucination detection beyond probability statistics.

\section*{Acknowledgments}
HM is supported by the Israel Science Foundation through
a personal grant (ISF 264/23) and an equipment grant
(ISF 532/23), and by the Career Advancement Chairs in
Artificial Intelligence -- Schmidt Futures.

G.B. is supported by the Jacobs Qualcomm PhD Fellowship.

F.F. conducted this work supported by an Aly Kaufman Post-Doctoral Fellowship.
    
\bibliographystyle{iclr2027_conference}    \bibliography{references}
    
    \clearpage
    \appendix
    
    \section{Additional Experimental Details}
    \label{app:experimental-details}
    
    \subsection{Datasets and Splits}
    \label{app:datasets-splits}
    
    For HotpotQA, we used the distractor configuration of
    \texttt{hotpotqa/hotpot\_qa}. We sampled 10,000 indices with replacement
    from each of its training and validation splits using seed 0;
    the latter served as our test set.
    For TriviaQA, we used \texttt{unfiltered-web-train.json} and
    \texttt{unfiltered-web-dev.json}, subsampling 10,000 examples from
    each with random state 42.
    
    For NQ-WC, we used the question, answer, and context fields of the
    LLMsKnow CSV. Removing rows with missing fields left 20,770 examples,
    which were partitioned into 10,000 training and 10,770 test examples
    with random state 42. Generation attempted 10,000 examples per split.
    The resulting data contained 9,674 training-pool examples and
    9,694 test examples. The 326 and 306 failed examples, respectively,
    were the examples with the longest contexts. Thus, the retained
    NQ-WC data excluded the longest-context cases.
    
    Detector training and validation used a label-stratified split
    constructed with \texttt{StratifiedKFold}, five folds, shuffling,
    and random state 42. Only the first fold was used for validation;
    the remaining folds were used for training. This gave an
    8,000/2,000 split for HotpotQA and TriviaQA.
    We did not perform five-fold cross-validation.
    Because correctness labels depended on the target model,
    the example-to-fold assignments could differ across models.
    
    
    \subsection{Prompts and Generation}
    \label{app:prompts-generation}
    
    The open-weight target models were
    \texttt{mistralai/Ministral-3-8B-Instruct-2512},
    \texttt{meta-llama/Llama-3.2-3B-Instruct}, and
    \texttt{google/gemma-4-12b-it}.
    Each prompt consisted of one user message without a system message,
    formatted using the target model's chat template with
    \texttt{add\_generation\_prompt=True}.
    HotpotQA and TriviaQA prompts contained only the question.
    NQ-WC prompts contained the context, a newline, and the question;
    the context was not truncated before generation.

    Generation used greedy decoding with
    \texttt{max\_new\_tokens=100} and \texttt{eos\_token\_id=None},
    which disabled stopping on end tokens. The models could still
    emit end tokens, but decoding continued to the 100-token cap.
    All stored responses in the nine model--dataset combinations
    contained exactly 100 generated tokens. Detector inputs used
    the full recorded sequences, including any continuation after
    the first end token.
    
    The generation pipeline retained full-vocabulary log-probabilities.
    Preprocessing retained the top $K=10$ candidates at each position.
    For detector input, candidate and actual-token log-probabilities
    were standardized using the mean and standard deviation of the
    finite log-probabilities over the full vocabulary at that position.
    Consequently, these inputs were standardized log-probabilities,
    rather than probabilities. This preprocessing required
    full-vocabulary statistics in the open-weight experiments.


    \subsection{Correctness Labels and Class Balance}
    \label{app:labels-balance}
    
    We followed the automatic answer-matching evaluation used in
    prior work \citep{HD-Guy,bar2026beyond}.
    A response received label 1 (factual) if its lowercased text
    contained a lowercased reference answer, and label 0
    (hallucination) otherwise. For TriviaQA, any official answer
    alias was accepted. HotpotQA and NQ-WC used a single reference
    string. No punctuation removal, whitespace normalization,
    or stemming was applied. The detector output represented
    $P(\text{factual})$.
    
    There was no separate refusal detector: a refusal received
    label 0 when it contained no reference match.
    Matching was performed over the complete generated response.
    In the open-weight experiments, this included the decoded text
    of all 100 generated tokens, including any continuation after
    the first end token. A reference match occurring only in this
    continuation was sufficient for a factual label.
    Short reference strings could also match within unrelated words.
    For context-based questions, repeating an answer string from
    the supplied passage could also receive credit without
    establishing that the response answered the question correctly.
    These labels therefore did not verify every claim in a response.
    
    The factual-label percentages in the stored data are shown below.
    ``Train pool'' denotes the data before the detector
    training/validation split.
    
    \begin{table}[t]
    \caption{Responses labeled factual (\%) in each dataset.
    Train pool denotes the data before the training/validation split.}
    \label{tab:class-balance}
    \centering
    \setlength{\tabcolsep}{5pt}
    \begin{tabular}{lrrrrrr}
    \toprule
    & \multicolumn{2}{c}{HotpotQA}
    & \multicolumn{2}{c}{TriviaQA}
    & \multicolumn{2}{c}{NQ-WC} \\
    Model & Train pool & Test & Train pool & Test & Train pool & Test \\
    \midrule
    Ministral & 33.60 & 29.98 & 60.99 & 60.87 & 54.77 & 54.62 \\
    Llama     & 28.45 & 24.97 & 58.32 & 60.23 & 60.74 & 60.52 \\
    Gemma     & 30.27 & 26.56 & 58.69 & 58.70 & 62.65 & 62.82 \\
    \bottomrule
    \end{tabular}
    \end{table}    
    
    \subsection{Detector Architecture and Training}
    \label{app:detector-training}
    
    GURI used two independent fusion MLPs, one shared across candidate
    slots and one for the actual-token branch. Each concatenated
    a 3,072-dimensional identity code with one standardized
    log-probability feature and applied
    \[
    \operatorname{Linear}(3073,384)
    \rightarrow \operatorname{ReLU}
    \rightarrow \operatorname{Dropout}(0.1)
    \rightarrow \operatorname{Linear}(384,1).
    \]
    The identity-code tables remained fixed.
    
    The response encoder used a hidden dimension of 256.
    The ten candidate features were projected to 128 dimensions;
    the actual-token feature and its rank supplied the other
    128 dimensions through learned scaling vectors.
    Rank encoding used $1-2r/V$, where $r$ was the zero-based
    actual-token rank and $V$ the target vocabulary size.
    The encoder used a learned classification token, learned
    positional embeddings, eight attention heads, a feedforward
    dimension of 256, and dropout 0.1. Classification used the
    classification-token representation followed by a linear
    layer and sigmoid.
    
    Training used AdamW, binary cross-entropy, batch size 48,
    and at most 100 epochs. The learning-rate schedule was linear
    with 10\% warm-up, with its length set from the planned
    100 epochs. Early stopping monitored validation ROC-AUC
    with patience 20.
    
    
    \subsection{Hyperparameter Search}
    \label{app:model-selection}
    
    The configured grid crossed training seeds $\{0,1,2\}$,
    encoder depths $\{2,3,4\}$, learning rates
    $\{10^{-4},5\times10^{-5}\}$, and weight decays
    $\{10^{-3},3\times10^{-3}\}$, giving 36 combinations.
    The hidden dimension, attention heads, dropout, batch size,
    and stopping rule were held fixed as specified in
    Appendix~\ref{app:detector-training}.
    Within each run, checkpoint selection used validation ROC-AUC. For the main comparison, we evaluated test ROC-AUC at each run's
    selected checkpoint and reported the mean and standard deviation
    across all 36 runs. These statistics therefore reflected variation
    across both hyperparameter configurations and training seeds.
    
    The channel-retraining experiments used a fixed configuration
    instead of this grid, as described in
    Appendix~\ref{app:channel-experiments}.
    
    
    \subsection{Identity-Code Tables}
    \label{app:identity-code-tables}
    
    Both code sources used the tokenizer of
    \texttt{microsoft/Phi-4-mini-instruct}.
    Each target token was decoded to a string using its source
    tokenizer and retokenized with the external tokenizer.
    
    For random codes, we sampled a 3,072-dimensional standard
    Gaussian vector for every external subtoken and normalized
    each vector to unit $\ell_2$ norm. We mean-pooled the vectors
    of a string's subtokens and normalized the pooled vector again.
    The generator seed was derived from the constant
    \texttt{PHI\_SCAFFOLD\_V1} using BLAKE2b with an eight-byte digest.
    The resulting table was shared across target models.
    Only one random codebook was used; sensitivity to different
    codebook draws was not evaluated.
    Tokens that could not be decoded or retokenized received
    the zero vector.
    
    For pretrained codes, we mean-pooled the corresponding
    3,072-dimensional Phi-4-mini input embeddings in float32.
    These pooled vectors were not $\ell_2$-normalized.
    The comparison therefore differed in both code source
    and normalization and did not isolate pretrained semantics.
    

    \subsection{GPT-4.1 Generation Setup}
    \label{app:gpt41-setup}
    
    We queried the Chat Completions API with model
    \texttt{gpt-4.1}, \texttt{temperature=0},
    \texttt{max\_tokens=50}, \texttt{logprobs=True}, and
    \texttt{top\_logprobs=10}.
    Each request contained one user message and no system message.
    Unlike the open-weight pipeline, the API supplied only the
    returned candidate log-probabilities rather than a
    full-vocabulary distribution. Missing candidate slots were
    padded with log-probability $-10^9$.
    The generation cap was 50 tokens, compared with 100 tokens
    in the open-weight experiments.

    \subsection{Alternating Transformer Architecture}
    \label{app:alternating-transformer}
    
    At each of $N$ generation positions, the model represented
    the $K=10$ candidates and a separate actual-token slot.
    For each slot, an MLP projected a fixed 3,072-dimensional code
(unit-norm random or unnormalized Phi-4-mini) to 254 dimensions,
using a hidden dimension of 384, ReLU, and dropout 0.1.
    Two features were then concatenated: the standardized
    log-probability and the probability.
    A learned indicator marked the actual-token slot.
    Layer normalization produced an $N\times(K+1)\times256$
    representation.
    
    Each alternating layer first applied attention across
    generation positions for each slot, using positional
    embeddings, and then across slots at each position.
    There were no candidate-position embeddings.
    After these layers, masked mean pooling over generation
    positions produced $K+1$ vectors. A further Transformer
    processed these vectors, followed by mean pooling over
    slots and an MLP with sigmoid output.
    The input codes remained fixed; the other components
    were trained jointly.
    
    The model used eight attention heads and dropout 0.1.
    The optimizer, batch size, schedule, epoch limit, and
    early-stopping rule followed
    Appendix~\ref{app:detector-training}.
    The search crossed the same depths, learning rates,
    weight decays, and training seeds listed in
    Appendix~\ref{app:model-selection}.
    
    
    

    \subsection{Channel Experiments}
    \label{app:channel-experiments}
    \label{sec:permutation-analysis}
    
    \paragraph{Retraining.}
    Each condition used three Transformer layers, hidden dimension
    256, eight attention heads, dropout 0.1, classification-token
    pooling, and scale rank encoding. Training used $K=10$,
    batch size 48, AdamW with learning rate $10^{-4}$ and
    weight decay $10^{-3}$, binary cross-entropy, and a linear
    schedule with 10\% warm-up. Runs lasted at most 100 epochs,
    with early stopping on validation ROC-AUC and patience 20.
    The train/validation split was the same as in the corresponding
    main experiment.
    
    We trained each condition with seeds 0, 1, and 2 and reported
    the mean and standard deviation of test ROC-AUC at each run's
    best-validation checkpoint. Evaluation used the full retained
    test set: 10,000 examples for HotpotQA and TriviaQA and
    9,694 for NQ-WC. The probability-only baseline was retrained
    with the same configuration and seeds.
    
    The full condition retained both identity channels.
    The actual-only condition replaced all non-actual candidate
    codes with zeros at data-loading time.
    The non-actual-only condition replaced the actual-token code
    with zeros in both its separate branch and every candidate
    slot with the same token ID. Token-ID matching identified
    these slots; the procedure did not assume a fixed rank.
    Probability features were unchanged.
    The random and pretrained code constructions followed
    Appendix~\ref{app:identity-code-tables}.
    
    \paragraph{Evaluation-time permutation.}
    For each model--dataset combination and code source, we selected
    the full-condition checkpoint with the highest validation
    ROC-AUC among the three training seeds and froze its parameters.
    We replaced the token-ID-to-code mapping with a derangement
    over the token IDs observed in the test split, so that no
    included ID retained its original assignment.
    
    We applied this mapping to actual-token entries, non-actual
    candidate entries, or all entries. Actual-token interventions
    included both the separate actual-token branch and matching
    candidate slots. All probability features remained unchanged.
    We evaluated the full test set with permutation seeds 0, 1,
    and 2 and reported the mean and standard deviation of the
    ROC-AUC decrease relative to the unpermuted evaluation of
    the same checkpoint. The Phi variant used the same procedure
    with pretrained codes computed on demand.
    
    These interventions measured the fitted detector's response
    to altered code assignments. They did not measure performance
    after retraining without the corresponding channel.
    
    
    
    \begin{figure}[H]
    \centering
    \begin{tikzpicture}
    
    \begin{groupplot}[
        group style={
            group size=3 by 1,
            horizontal sep=1.1cm,
            ylabels at=edge left
        },
        width=0.335\textwidth,
        height=5.3cm,
        ybar=1.5pt,
        /pgf/bar width=7pt,
        xmin=0.5,
        xmax=3.5,
        xtick={1,2,3},
        xticklabels={Gemma,Llama,Ministral},
        tick label style={font=\scriptsize},
        title style={font=\small\bfseries},
        ylabel style={font=\small},
        ymajorgrids=true,
        grid style={gray!20},
        axis line style={gray!70},
        legend style={
            font=\scriptsize,
            draw=none,
            legend columns=3,
            column sep=7pt
        },
        legend image code/.code={\fill[#1] (0cm,-0.1cm) rectangle (0.25cm,0.1cm);}
    ]
    
    \nextgroupplot[
        title={TriviaQA},
        ymin=0,
        ymax=30,
        ytick={0,10,20,30},
        ylabel={Test ROC-AUC lost (points)},
        legend to name=identitychannelslegend
    ]
    
    \addplot+[
        ybar,
        fill=embcolor!60,
        draw=embcolor,
        error bars/.cd,
        y dir=both,
        y explicit,
        error bar style={gray!60!black}
    ] coordinates {
        (1,14.7) +- (0,2.3)
        (2,24.6) +- (0,4.1)
        (3,12.5) +- (0,1.2)
    };
    \addlegendentry{all tokens}
    
    \addplot+[
        ybar,
        fill=myblue!70,
        draw=myblue,
        error bars/.cd,
        y dir=both,
        y explicit,
        error bar style={gray!60!black}
    ] coordinates {
        (1,5.2) +- (0,2.0)
        (2,6.8) +- (0,1.9)
        (3,10.1) +- (0,0.8)
    };
    \addlegendentry{actual token only}
    
    \addplot+[
        ybar,
        fill=extteal!70,
        draw=extteal,
        error bars/.cd,
        y dir=both,
        y explicit,
        error bar style={gray!60!black}
    ] coordinates {
        (1,5.7) +- (0,0.1)
        (2,1.8) +- (0,0.1)
        (3,0.5) +- (0,0.1)
    };
    \addlegendentry{non-actual candidates only}
    
    \nextgroupplot[
        title={HotpotQA},
        ymin=0,
        ymax=50,
        ytick={0,10,20,30,40,50}
    ]
    
    \addplot+[
        ybar,
        fill=embcolor!60,
        draw=embcolor,
        error bars/.cd,
        y dir=both,
        y explicit,
        error bar style={gray!60!black}
    ] coordinates {
        (1,24.8) +- (0,4.9)
        (2,38.5) +- (0,9.7)
        (3,13.6) +- (0,2.3)
    };
    
    \addplot+[
        ybar,
        fill=myblue!70,
        draw=myblue,
        error bars/.cd,
        y dir=both,
        y explicit,
        error bar style={gray!60!black}
    ] coordinates {
        (1,9.6) +- (0,1.2)
        (2,7.7) +- (0,3.8)
        (3,8.3) +- (0,1.6)
    };
    
    \addplot+[
        ybar,
        fill=extteal!70,
        draw=extteal,
        error bars/.cd,
        y dir=both,
        y explicit,
        error bar style={gray!60!black}
    ] coordinates {
        (1,4.3) +- (0,0.6)
        (2,5.8) +- (0,0.2)
        (3,1.3) +- (0,0.2)
    };
    
    \nextgroupplot[
        title={NQ (with context)},
        ymin=0,
        ymax=40,
        ytick={0,10,20,30,40}
    ]
    
    \addplot+[
        ybar,
        fill=embcolor!60,
        draw=embcolor,
        error bars/.cd,
        y dir=both,
        y explicit,
        error bar style={gray!60!black}
    ] coordinates {
        (1,29.4) +- (0,5.6)
        (2,21.3) +- (0,2.2)
        (3,21.1) +- (0,1.2)
    };
    
    \addplot+[
        ybar,
        fill=myblue!70,
        draw=myblue,
        error bars/.cd,
        y dir=both,
        y explicit,
        error bar style={gray!60!black}
    ] coordinates {
        (1,10.8) +- (0,3.1)
        (2,11.8) +- (0,1.7)
        (3,9.7) +- (0,1.3)
    };
    
    \addplot+[
        ybar,
        fill=extteal!70,
        draw=extteal,
        error bars/.cd,
        y dir=both,
        y explicit,
        error bar style={gray!60!black}
    ] coordinates {
        (1,5.8) +- (0,0.1)
        (2,2.0) +- (0,0.1)
        (3,2.7) +- (0,0.2)
    };
    
    \end{groupplot}
    
    \node[anchor=north] at
    ($(group c1r1.south)!0.5!(group c3r1.south)+(0,-0.8cm)$)
    {\pgfplotslegendfromname{identitychannelslegend}};
    
    \end{tikzpicture}
    
    \caption{
    \textbf{The fitted detector relies on both identity channels.}
    Test ROC-AUC lost when, with the trained GURI frozen, the token-to-code correspondence is permuted at
    evaluation time (a derangement of all token identities) in every slot, in the actual-token channel only (its
    ATP entry and its candidate slot at every step), or in the non-actual candidates' slots only. Unit-norm random
    codes, three LLMs, three datasets, bars are the mean over 3 permutation seeds and error bars one standard
    deviation. Both channels cost AUC in every cell, and the joint loss exceeds the sum of the two everywhere, by a
    wide margin in most cells: neither channel is read in isolation. The fitted detector leans more on the actual
    token (the larger loss in eight of nine cells), yet a permutation loss measures what this detector depends on,
    not what a channel is worth: retrained without one channel (Fig.~\ref{fig:channel-retraining}), the non-actual
    candidates alone match the full model while the actual token alone does not.
    }
    \label{fig:identity-channels}
    \end{figure}
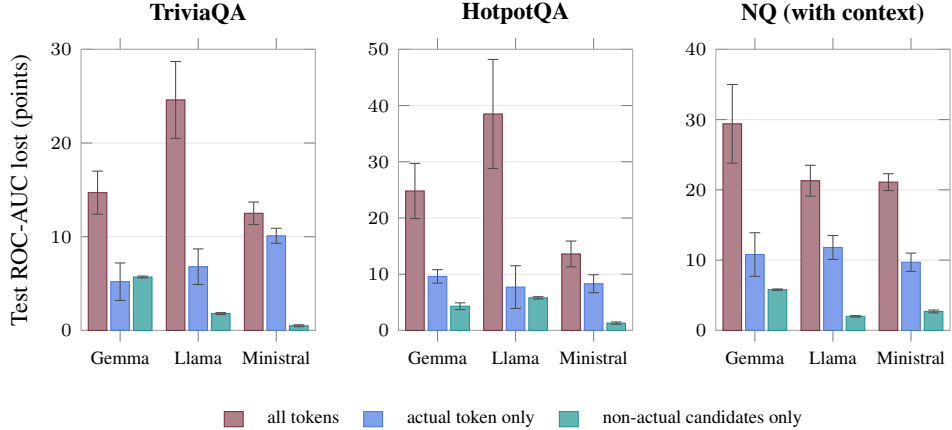
    
    Figure~\ref{fig:identity-channels} shows that performance decreases when either
    channel is perturbed: the fitted detector relies on both actual-token and
    non-actual candidate identities. The substantially larger losses under joint
    permutation are non-additive and do not by themselves establish an interaction
    between the two channels. Moreover, this experiment measures reliance by a
    fixed detector; it does not determine how well a model can perform when trained
    without one of the channels.
    
    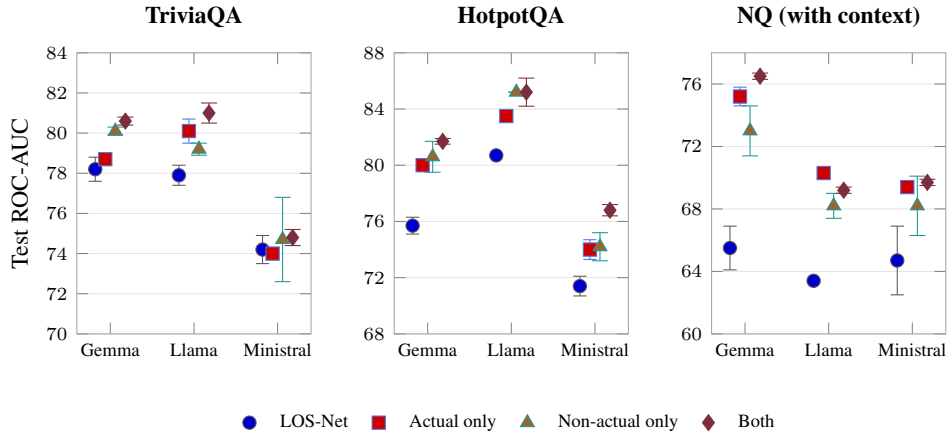
\begin{figure*}[t]
    \centering
    \begin{tikzpicture}
    
    \begin{groupplot}[
        group style={
            group size=3 by 1,
            horizontal sep=1.1cm,
            ylabels at=edge left
        },
        width=0.335\textwidth,
        height=5.3cm,
        xmin=0.6,
        xmax=3.4,
        xtick={1,2,3},
        xticklabels={Gemma,Llama,Ministral},
        tick label style={font=\scriptsize},
        title style={font=\small\bfseries},
        ylabel style={font=\small},
        ymajorgrids=true,
        grid style={gray!20},
        axis line style={gray!70},
        legend style={
            font=\scriptsize,
            draw=none,
            legend columns=4,
            column sep=7pt
        }
    ]
    
    \nextgroupplot[
        title={TriviaQA},
        ymin=70,
        ymax=84,
        ytick={70,72,74,76,78,80,82,84},
        ylabel={Test ROC-AUC},
        legend to name=channelretrainingphilegend
    ]
    
    \addplot+[
        only marks,
        mark=*,
        mark size=2.5pt,
        color=gray!70!black,
        error bars/.cd,
        y dir=both,
        y explicit
    ] coordinates {
        (0.82,78.2) +- (0,0.6)
        (1.82,77.9) +- (0,0.5)
        (2.82,74.2) +- (0,0.7)
    };
    \addlegendentry{LOS-Net}
    
    \addplot+[
        only marks,
        mark=square*,
        mark size=2.5pt,
        color=myblue,
        error bars/.cd,
        y dir=both,
        y explicit
    ] coordinates {
        (0.94,78.7) +- (0,0.1)
        (1.94,80.1) +- (0,0.6)
        (2.94,74.0) +- (0,0.3)
    };
    \addlegendentry{Actual only}
    
    \addplot+[
        only marks,
        mark=triangle*,
        mark size=3pt,
        color=extteal,
        error bars/.cd,
        y dir=both,
        y explicit
    ] coordinates {
        (1.06,80.1) +- (0,0.2)
        (2.06,79.2) +- (0,0.3)
        (3.06,74.7) +- (0,2.1)
    };
    \addlegendentry{Non-actual only}
    
    \addplot+[
        only marks,
        mark=diamond*,
        mark size=3pt,
        color=embcolor,
        error bars/.cd,
        y dir=both,
        y explicit
    ] coordinates {
        (1.18,80.6) +- (0,0.2)
        (2.18,81.0) +- (0,0.5)
        (3.18,74.8) +- (0,0.4)
    };
    \addlegendentry{Both}
    
    \nextgroupplot[
        title={HotpotQA},
        ymin=68,
        ymax=88,
        ytick={68,72,76,80,84,88}
    ]
    
    \addplot+[
        only marks,
        mark=*,
        mark size=2.5pt,
        color=gray!70!black,
        error bars/.cd,
        y dir=both,
        y explicit
    ] coordinates {
        (0.82,75.7) +- (0,0.6)
        (1.82,80.7) +- (0,0.0)
        (2.82,71.4) +- (0,0.7)
    };
    
    \addplot+[
        only marks,
        mark=square*,
        mark size=2.5pt,
        color=myblue,
        error bars/.cd,
        y dir=both,
        y explicit
    ] coordinates {
        (0.94,80.0) +- (0,0.4)
        (1.94,83.5) +- (0,0.4)
        (2.94,74.0) +- (0,0.7)
    };
    
    \addplot+[
        only marks,
        mark=triangle*,
        mark size=3pt,
        color=extteal,
        error bars/.cd,
        y dir=both,
        y explicit
    ] coordinates {
        (1.06,80.6) +- (0,1.1)
        (2.06,85.2) +- (0,0.0)
        (3.06,74.2) +- (0,1.0)
    };
    
    \addplot+[
        only marks,
        mark=diamond*,
        mark size=3pt,
        color=embcolor,
        error bars/.cd,
        y dir=both,
        y explicit
    ] coordinates {
        (1.18,81.7) +- (0,0.2)
        (2.18,85.2) +- (0,1.0)
        (3.18,76.8) +- (0,0.4)
    };
    
    \nextgroupplot[
        title={NQ (with context)},
        ymin=60,
        ymax=78,
        ytick={60,64,68,72,76}
    ]
    
    \addplot+[
        only marks,
        mark=*,
        mark size=2.5pt,
        color=gray!70!black,
        error bars/.cd,
        y dir=both,
        y explicit
    ] coordinates {
        (0.82,65.5) +- (0,1.4)
        (1.82,63.4) +- (0,0.1)
        (2.82,64.7) +- (0,2.2)
    };
    
    \addplot+[
        only marks,
        mark=square*,
        mark size=2.5pt,
        color=myblue,
        error bars/.cd,
        y dir=both,
        y explicit
    ] coordinates {
        (0.94,75.2) +- (0,0.6)
        (1.94,70.3) +- (0,0.4)
        (2.94,69.4) +- (0,0.4)
    };
    
    \addplot+[
        only marks,
        mark=triangle*,
        mark size=3pt,
        color=extteal,
        error bars/.cd,
        y dir=both,
        y explicit
    ] coordinates {
        (1.06,73.0) +- (0,1.6)
        (2.06,68.2) +- (0,0.8)
        (3.06,68.2) +- (0,1.9)
    };
    
    \addplot+[
        only marks,
        mark=diamond*,
        mark size=3pt,
        color=embcolor,
        error bars/.cd,
        y dir=both,
        y explicit
    ] coordinates {
        (1.18,76.5) +- (0,0.2)
        (2.18,69.2) +- (0,0.2)
        (3.18,69.7) +- (0,0.2)
    };
    
    \end{groupplot}
    
    \node[anchor=north] at
    ($(group c1r1.south)!0.5!(group c3r1.south)+(0,-0.8cm)$)
    {\ref{channelretrainingphilegend}};
    
    \end{tikzpicture}
    
    \caption{
    \textbf{The same retraining with pretrained Phi-4-mini embeddings in place of random codes.}
    Layout and hyper-parameters as in Fig.~\ref{fig:channel-retraining}; each token's vector is the mean of the
    Phi-4-mini input embeddings of its sub-tokens. The non-actual-only detector is lower than with random codes
    in every cell (by 2 to 7 AUC points), so the full model now needs the actual-token channel to reach its best
    value, and on NQ the actual token alone is worth more than the non-actual candidates. The full model itself
    is below its random-code counterpart in all nine cells. Mean over 3 training seeds, error bars one
    standard deviation.
    }
    \label{fig:channel-retraining-phi}
    \end{figure*}

    \subsection{Transfer Evaluation Details}
    \label{app:cross-model-transfer}
    \label{app:cross-dataset-transfer}

We evaluated LOS-Net and GURI with unit-norm random codes across
three models (Gemma, Llama, and Ministral) and three
datasets (NQ-WC, HotpotQA, and TriviaQA).

For each method and source model--dataset pair, we selected the
checkpoint with the highest validation ROC-AUC among its 36 training
runs. We evaluated that checkpoint on all nine target pairs without
fine-tuning or target-validation selection. Input features, labels,
and identity codes came from the target data. GURI used the shared
random code table described in
Appendix~\ref{app:identity-code-tables}.
Evaluation used 10,000 test examples for HotpotQA and TriviaQA
and 9,694 for NQ-WC.

Each method had 81 evaluations: nine without a shift, 18 changing
only the model, 18 changing only the dataset, and 36 changing both.
Each score came from one selected checkpoint, so diagonal values
need not equal the averages in Table~\ref{tab:main-results}.
    
\begin{figure}[p]
\centering
\begingroup
\definecolor{transferPositive}{RGB}{217,130,88}
\definecolor{transferNegative}{RGB}{115,145,201}

\def\transferConfidence{{61.13,60.94,64.98,67.29,55.08,64.40,55.29,53.27,59.28}}
\def\transferTargetLOS{{77.93,82.05,70.30,78.74,79.17,74.29,68.27,63.98,66.98}}

\newcommand{\transferRow}[2]{%
\foreach \HeatAUC [count=\HeatCol from 0] in {#2}{
  \pgfmathsetmacro{\HeatGain}{
    \HeatAUC-\transferConfidence[\HeatCol]
  }
  \pgfmathsetmacro{\HeatTint}{
    min(100,abs(\HeatGain)/30*100)
  }
  \ifdim\HeatGain pt<0pt
    \fill[transferNegative!\HeatTint!white]
      (\HeatCol,{-#1-1}) rectangle ++(1,1);
  \else
    \fill[transferPositive!\HeatTint!white]
      (\HeatCol,{-#1-1}) rectangle ++(1,1);
  \fi
  \draw[white,line width=0.5pt]
    (\HeatCol,{-#1-1}) rectangle ++(1,1);
  \node[font=\scriptsize\bfseries\boldmath]
    at ({\HeatCol+0.5},{-#1-0.32})
    {\pgfmathprintnumber[
      fixed,precision=2,zerofill
    ]{\HeatAUC}};
  \node[font=\tiny]
    at ({\HeatCol+0.5},{-#1-0.73})
    {\pgfmathprintnumber[
      fixed,precision=1,zerofill,showpos
    ]{\HeatGain}};
  \ifnum\HeatStars=1
    \pgfmathtruncatemacro{\HeatMarked}{
      \HeatAUC>\transferTargetLOS[\HeatCol]
    }
    \ifnum\HeatMarked=1
      \draw[black,line width=0.5pt]
        ({\HeatCol+0.035},{-#1-0.965})
        rectangle ++(0.93,0.93);
      \node[font=\tiny,inner sep=0pt]
        at ({\HeatCol+0.89},{-#1-0.16}) {$\star$};
    \fi
  \fi
}}

\newcommand{\transferAxisEntry}[3]{%
  \node[font=\tiny]
    at ({#1+0.5},0.95) {#2};
  \node[anchor=east,font=\scriptsize]
    at (-0.12,{-#1-0.5}) {#2};
  \node[font=\scriptsize]
    at ({#1+0.5},0.35) {#3};
}

\newcommand{\transferAxes}[1]{%
  \node[font=\bfseries] at (4.5,2.05) {#1};
  \node[font=\scriptsize\bfseries,text=transferPositive]
    at (3,1.70) {closed-book};
  \node[font=\scriptsize\bfseries,text=transferNegative]
    at (7.5,1.70) {open-book};
  \draw[black,line width=0.8pt]
    (0.08,1.25)
    .. controls (0.08,1.37) and (0.18,1.40) .. (0.26,1.40)
    -- (2.77,1.40)
    .. controls (2.88,1.40) and (2.93,1.31) .. (3,1.25)
    .. controls (3.07,1.31) and (3.12,1.40) .. (3.23,1.40)
    -- (5.74,1.40)
    .. controls (5.82,1.40) and (5.92,1.37) .. (5.92,1.25);
  \draw[black,line width=0.8pt]
    (6.08,1.25)
    .. controls (6.08,1.37) and (6.18,1.40) .. (6.26,1.40)
    -- (7.27,1.40)
    .. controls (7.38,1.40) and (7.43,1.31) .. (7.5,1.25)
    .. controls (7.57,1.31) and (7.62,1.40) .. (7.73,1.40)
    -- (8.74,1.40)
    .. controls (8.82,1.40) and (8.92,1.37) .. (8.92,1.25);

  \transferAxisEntry{0}{G/H}{61.13}
  \transferAxisEntry{1}{L/H}{60.94}
  \transferAxisEntry{2}{M/H}{64.98}
  \transferAxisEntry{3}{G/T}{67.29}
  \transferAxisEntry{4}{L/T}{55.08}
  \transferAxisEntry{5}{M/T}{64.40}
  \transferAxisEntry{6}{G/N}{55.29}
  \transferAxisEntry{7}{L/N}{53.27}
  \transferAxisEntry{8}{M/N}{59.28}

  \node[anchor=east,font=\scriptsize]
    at (-0.12,0.95) {Target};
  \node[anchor=east,font=\scriptsize]
    at (-0.12,0.35) {Confidence};
  \node[rotate=90,font=\small]
    at (-0.9,-4.5) {Source};

  \draw[black!60,line width=0.6pt] (3,0) -- (3,-9);
  \draw[black!80,line width=1.1pt] (6,0) -- (6,-9);
  \draw[black!60,line width=0.6pt] (0,-3) -- (9,-3);
  \draw[black!80,line width=1.1pt] (0,-6) -- (9,-6);
  \draw[black,line width=0.6pt] (0,0) rectangle (9,-9);
}

\resizebox{0.95\linewidth}{!}{%
\begin{tikzpicture}[x=1.1cm,y=0.60cm]

\begin{scope}
\def\HeatStars{0}
\transferRow{0}{77.93,54.09,55.99,74.94,67.55,59.63,50.41,51.07,52.82}
\transferRow{1}{43.64,82.05,59.74,39.23,70.29,52.28,47.44,56.64,52.97}
\transferRow{2}{62.29,59.55,70.30,67.98,64.24,70.16,49.67,45.77,48.79}
\transferRow{3}{68.31,48.78,55.94,78.74,57.29,59.25,50.27,49.44,56.95}
\transferRow{4}{59.35,67.87,56.30,60.90,79.17,57.88,50.27,60.15,48.40}
\transferRow{5}{62.91,51.91,65.73,68.96,58.42,74.29,49.83,47.16,52.02}
\transferRow{6}{49.07,57.35,53.12,41.82,54.04,50.00,68.27,53.60,51.73}
\transferRow{7}{55.62,58.29,53.14,54.11,62.15,48.49,57.76,63.98,53.02}
\transferRow{8}{61.62,36.87,54.17,62.15,49.15,51.85,53.28,58.32,66.98}
\transferAxes{LOS-Net}
\end{scope}

\begin{scope}[yshift=-7.4cm]
\def\HeatStars{1}
\transferRow{0}{84.84,64.43,74.98,77.51,61.42,67.98,54.13,52.27,48.80}
\transferRow{1}{74.88,87.31,68.87,68.01,76.67,58.18,55.31,54.91,52.00}
\transferRow{2}{73.99,49.74,78.77,73.30,43.22,70.28,59.36,55.18,49.99}
\transferRow{3}{72.44,54.25,62.87,82.69,69.75,69.30,53.48,51.13,44.99}
\transferRow{4}{65.12,76.55,61.93,73.37,81.97,68.26,58.42,58.07,47.44}
\transferRow{5}{66.11,62.94,66.41,76.53,73.12,77.56,53.76,51.98,51.04}
\transferRow{6}{58.72,42.09,47.48,54.41,54.50,45.79,79.00,70.41,67.85}
\transferRow{7}{56.70,45.63,46.87,55.64,55.80,48.05,67.41,73.33,65.62}
\transferRow{8}{51.38,40.87,47.72,52.92,50.79,51.37,69.28,68.26,73.96}
\transferAxes{\raisebox{2pt}[0pt][0pt]{GURI: unit-norm random codes}}\end{scope}

\begin{scope}[yshift=-14cm]
\foreach \i in {0,...,60}{%
  \pgfmathtruncatemacro{\legendGain}{\i-30}
  \pgfmathsetmacro{\legendTint}{abs(\legendGain)/30*100}
  \pgfmathsetmacro{\legendX}{2+5*\i/61}
  \ifdim\legendGain pt<0pt
    \fill[transferNegative!\legendTint!white]
      (\legendX,0) rectangle ++({5/61+0.002},0.3);
  \else
    \fill[transferPositive!\legendTint!white]
      (\legendX,0) rectangle ++({5/61+0.002},0.3);
  \fi
}
\draw[line width=0.4pt]
  (2,0) rectangle (7,0.3);
\foreach \HeatX/\HeatLabel in {
  2/-30,3.25/-15,4.5/0,5.75/+15,7/+30
}{
  \node[anchor=north,font=\scriptsize]
    at (\HeatX,-0.05) {\HeatLabel};
}
\node[font=\scriptsize] at (4.5,-1)
  {Gain over target Confidence (AUC points)};
\end{scope}

\end{tikzpicture}%
}
\endgroup
\caption{Complete transfer results. Rows and columns denote source and
target model--dataset pairs: G, L, M denote Gemma, Llama, Ministral.
H and T are closed-book; N denotes open-book NQ-WC.
Bold values show test ROC-AUC (\%); signed values and colors show
differences from target Confidence, listed above the columns.
Stars and outlines mark GURI scores above target-trained LOS-Net,
not statistical significance. Diagonal dataset blocks correspond to the
within-dataset cross-model panels in Figure~\ref{fig:cross-model-transfer}.
Each row uses one source-validation-selected checkpoint without
target fine-tuning.}
\label{tab:cross-model-transfer}
\end{figure}
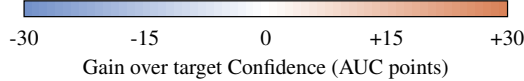

    Table~\ref{tab:dataset-transfer-summary} reports unweighted means
    over the dataset-shift conditions.
    HotpotQA and TriviaQA supplied only the question, whereas NQ-WC
    also supplied context. Thus, transfers involving NQ-WC changed
    both the dataset and context availability.
    Figure~\ref{tab:cross-model-transfer} reports all individual scores.
    
    \begin{table}[t]
    \caption{Mean test ROC-AUC (\%) under dataset shifts.
    $H$, $T$, and $N$ denote HotpotQA, TriviaQA, and NQ-WC.
    ``Same'' and ``Different'' indicate whether the source and target
    models coincide; $n$ is the number of directed transfers.
   Confidence uses the target-specific baseline scores, averaged
over the same target cells.
    Means summarize transfer settings, not repeated training runs.}
    \label{tab:dataset-transfer-summary}
    \centering
    \setlength{\tabcolsep}{5pt}\begin{tabular}{llrrrr}
    \toprule
    Dataset transfer & Model & $n$ & Confidence & LOS-Net & GURI \\
    \midrule
$H \leftrightarrow T$
& Same      & 6  & 62.30 & 69.55 & 73.31 \\
$H \leftrightarrow T$
& Different & 12 & 62.30 & 57.18 & 62.11 \\
$N \leftrightarrow \{H,T\}$
& Same      & 12 & 59.13 & 52.97 & 52.94 \\
$N \leftrightarrow \{H,T\}$
& Different & 24 & 59.13 & 51.56 & 50.99 \\
    \bottomrule
    \end{tabular}
    \end{table}

    For the HotpotQA transfer from Ministral to Llama, GURI achieved
    49.74 ROC-AUC, compared with 68.87 in the reverse direction.
    A check confirmed that the low-scoring evaluation loaded the
    3,072-dimensional target identity codes. This ruled out missing
    identity codes as the explanation, but did not explain the result.

    \subsection{Compute}
    \label{app:compute}
    
    Experiments were run on an NVIDIA DGX system equipped with
    eight A100-SXM4 GPUs with 40\,GB memory each, 256 CPU cores,
    and 1,007\,GB system memory.
    The system used NVIDIA driver 555.42.06 and CUDA 12.5.

    \end{document}